\documentclass{article}

\usepackage{booktabs}
\usepackage{multirow}

\usepackage[final]{corl_2026}

\usepackage{enumitem}
\usepackage{amsmath}
\usepackage{amssymb}
\usepackage{algorithm}
\usepackage{algpseudocode}
\usepackage{adjustbox}
\usepackage{graphicx}
\usepackage{subcaption}
\usepackage{tabularx}
\usepackage{array}
\usepackage{wrapfig}

\graphicspath{{figures/}}

\title{ESCAPE: Egocentric Sensing for Collision Avoidance and Policy Execution}

\author{
{\bfseries
Weiqing Shen$^{*}$ \quad
Wenhao Huang$^{*}$ \quad
Baokang Zhang \quad
Shuo Shang
} \\
{\bfseries
Yanmiao Hu \quad
Haojun Li \quad
Yifan Han \quad
Wenzhao Lian$^{\dagger}$
} \\[2mm]
Shanghai Jiao Tong University\\[1mm]
$^{*}$Equal contribution \qquad
$^{\dagger}$Corresponding author
}

\begin{document}

\maketitle


\begin{figure}[h] 
    \centering
    \includegraphics[width=0.92\textwidth]{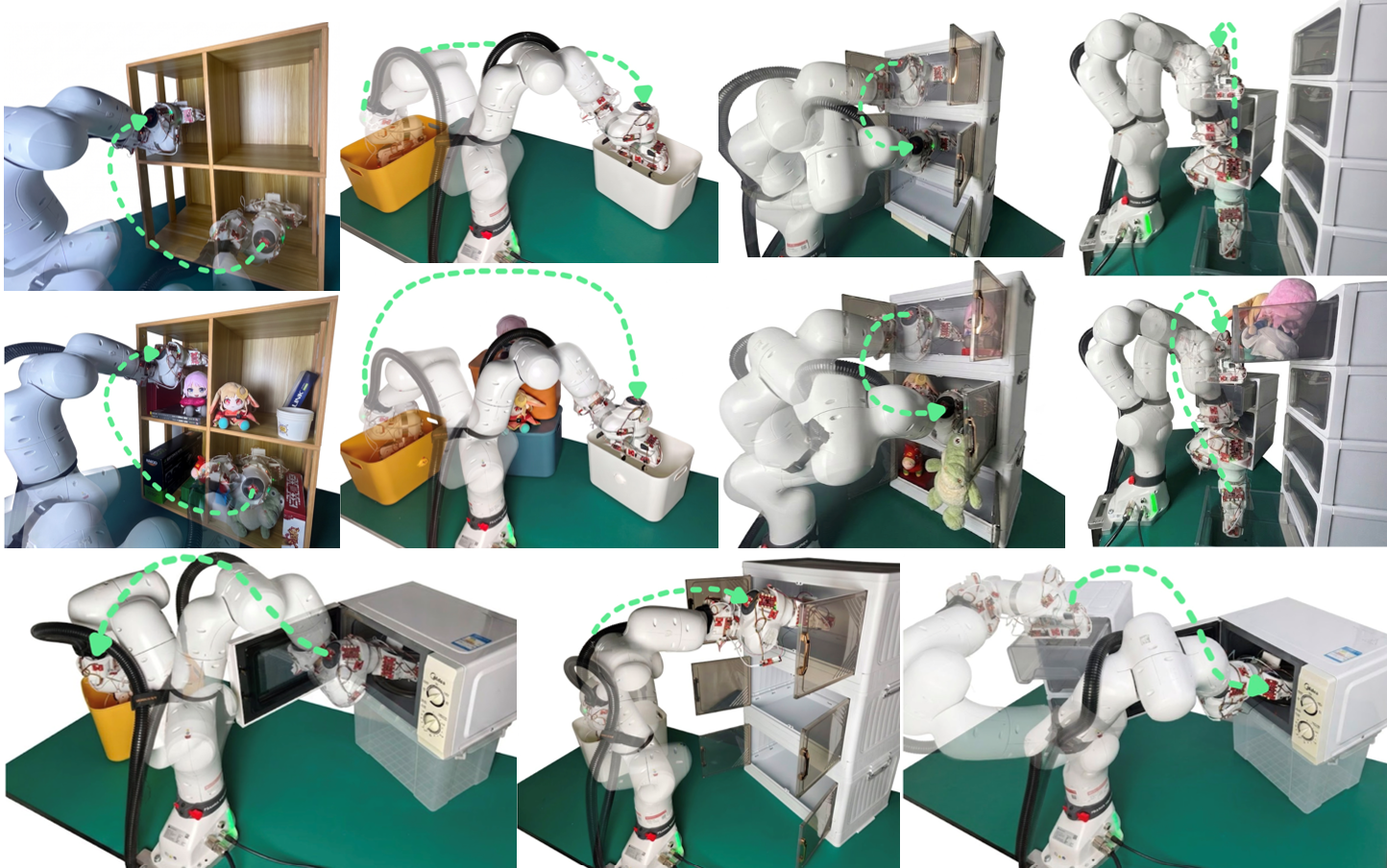} 
    \caption{We introduce ESCAPE, an online egocentric sensing framework capable of generating safe, collision-free trajectories in highly cluttered and unknown environments.}
    \label{fig:figure1}
\end{figure}
\begin{abstract}
Motion planning in unstructured environments is a bottleneck for robot deployment. Current visuomotor policies predominantly rely on external cameras, introducing perceptual blind spots and fragility due to occlusions, clutter, and calibration constraints. To overcome these limitations, we propose ESCAPE, an online egocentric sensing framework for collision-free trajectory generation. ESCAPE aggregates sparse measurements from distributed depth sensors, converting them into a structured, obstacle-aware point cloud representation. Our policy leverages a flow matching-based generative framework, and trains on a high-quality dataset comprising over 100K safety-critical trajectories. With joint optimization of goal-reaching and collision objectives, it produces safe trajectories directly from these egocentric inputs. Crucially, ESCAPE is trained entirely in simulation and transfers zero-shot to physical hardware without any real-world fine-tuning. Extensive simulation and real-world evaluations demonstrate that ESCAPE achieves stable motion planning using only egocentric perception, outperforming external-view and voxelized-planning baselines across both simulation and real-world evaluations. Our website: \href{https://escape-policy.github.io/}{escape-policy.github.io/}.
\end{abstract}



\section{Introduction}
Autonomous systems operating in the wild, such as self-driving vehicles, rely fundamentally on onboard, egocentric perception to navigate in complex environments. By coupling the perceptual frames directly to the agent's motion, these systems maintain situational awareness independent of pre-installed external infrastructure.
In contrast, many robot motion-generation systems still rely on third-person cameras or globally reconstructed point clouds, which can be fragile in cluttered manipulation settings due to self-occlusion, scene occlusion, camera displacement, and careful placement or calibration requirements. 

To bypass these limitations, modern paradigms have shifted away from classical motion planning methods~\citep{orthey2023sampling, 4082128, karaman2011sampling, strub2020adaptively} -- which generate reliable trajectories but require accurate environment geometric models --
and moved towards learning based motion policies. These learning-based policies distill expert planning experience into perception-conditioned action generation. 
Recent end-to-end motion planning methods have widely adopted point clouds to improve geometric generalization in cluttered environments~\citep{fishman2023motion, pmlr-v270-fishman25a, dalal2024neural, yang2025deep}. 
However, these methods still predominantly depend on external-view observations or global point-cloud reconstructions, which can be difficult to maintain in real-world workspaces with occlusions and changing camera configurations.

Egocentric sensing offers a promising direction by mounting sensors directly on the robot body. Moving with the robot, they capture near-field geometry and reduce reliance on fixed external cameras. While recent work like ARMOR~\citep{kim2024armor} demonstrates the potential of body-mounted ToF sensors for collision avoidance, cluttered manipulator reaching remains challenging due to the local, motion-dependent, and highly sparse nature of these observations.

To address this gap, we propose ESCAPE, an online egocentric motion-generation framework for collision-free reaching in cluttered and partially observed environments. ESCAPE deploys distributed body-mounted depth sensors and generates safe motion trajectories from egocentric sensory inputs, robot state, and goal information. 
To handle sparse near-field observations, we temporally aggregate multi-joint ToF measurements and convert them into an obstacle-avoidance-aware point-cloud representation. 
We then use a flow matching-based generative policy, trained from collision-free expert trajectories generated by cuRobo~\citep{sundaralingam2023curobo}, and jointly optimize goal-reaching and collision objectives to guide safe trajectory generation. 
For controlled comparison with prior work, our simulation pipeline uses scene categories adapted from DRP~\citep{yang2025deep}; training on broader environment distributions is left for future work.
Importantly, once the policy is trained in simulation, 
the same robot-mounted egocentric sensing setup can be deployed in the real world without external camera placement, multi-camera calibration, or scene-dependent view adjustment.
As validated in our experiments, ESCAPE is deployed zero-shot onto a real Franka robot, demonstrating a plug-and-play sim-to-real path for collision-aware manipulation.

Our core contributions are:
\begin{itemize}[leftmargin=2em]
    \item We present an online egocentric motion-generation framework that uses robot-mounted near-field depth sensing to generate safe, goal-directed trajectories in cluttered environments.
    \item We design a temporal aggregation representation to convert sparse distributed ToF measurements into a structured, obstacle-aware point cloud for collision-aware policy learning.
    \item We build a simulation-only training pipeline with obstacle-proximal expert demonstrations and synchronized egocentric observations, producing over 100K safety-critical trajectories.
    \item We demonstrate zero-shot sim-to-real deployment on a real Franka robot, showing that egocentric sensing avoids careful external-camera placement, calibration, and view adjustment.
\end{itemize}

Extensive evaluations in cluttered environments show that ESCAPE outperforms external-view and voxelized-planning baselines using only egocentric perception, achieving robust zero-shot deployment on a real robot.



\section{Related Work}

\subsection{Classical Motion Planning}

Classical motion planning generates collision-free trajectories by searching or optimizing over the robot configuration space. Graph-search methods such as A* and its variants~\citep{4082128, likhachev2003ara, likhachev2005anytime, koenig2006new} provide strong search guarantees under appropriate assumptions, while sampling-based planners such as RRT, PRM, and their extensions~\citep{trees1998new, kavraki2002probabilistic, karaman2011sampling, strub2020adaptively} improve scalability in high-dimensional continuous spaces. Trajectory optimization methods further refine paths by minimizing smoothness, goal, and collision costs~\citep{zucker2013chomp, dragan2011manipulation, schulman2014motion, sundaralingam2023curobo}. Control Barrier Function-based methods provide another line of safety-aware planning and control~\citep{singletary2022safety, thirugnanam2022safety, yu2024efficient}. Despite strong performance with accurate geometric models, these approaches typically depend on explicit environment representations and collision queries, making them less suitable for unknown, cluttered, or severely occluded real-world scenes.

\subsection{Learning-based Motion Planning}

Learning-based motion planners aim to improve generalization and responsiveness by distilling planning experience into neural policies. Early methods use learned modules to bias sampling, generate roadmaps, or guide classical planners~\citep{qureshi2019motion,kumar2019lego,zhang2018learning,chamzas2021learning,ichter2020broadly,qureshi2020neural}. More recent closed-loop visuomotor policies directly map partial point-cloud observations and robot states to actions. M$\pi$Nets~\citep{fishman2023motion} and M$\pi$Former~\citep{pmlr-v270-fishman25a} demonstrate online neural motion generation from point-cloud inputs, while NeuralMP~\citep{dalal2024neural} improves generalization through large-scale training and test-time optimization. DRP~\citep{yang2025deep} further enhances reactive manipulator planning in dynamic environments by combining a pretrained motion policy with local reactive modules. However, these methods still primarily rely on external cameras or global point-cloud observations. In cluttered workspaces, such observations can be degraded by self-occlusion, scene occlusion, and camera-placement constraints, motivating alternative sensing strategies that are directly coupled to robot motion.

\subsection{Egocentric Perception}

Egocentric perception places sensors on the robot body or near the end effector, providing local observations that move with the robot. This paradigm has been explored in visual manipulation through whole-body vision, active egocentric cameras, and in-hand view reasoning~\citep{xu2025robopanoptes, ding2025imagination, zou2026activeglasses}. For collision avoidance, proximity skins and distributed depth sensors provide near-field measurements without requiring a complete scene model~\citep{tsuji2019proximity,giovinazzo2024cyskin,schlegl2013virtual}. ARMOR~\citep{kim2024armor} recently integrates multiple ToF sensors on humanoid arms and demonstrates the value of body-centric perception for collision avoidance and motion planning. However, achieving robust performance using purely body-mounted perception on real physical robots remains a significant challenge, and existing methods, including ARMOR, have yet to demonstrate stable and highly effective results in complex, real-world cluttered environments.

\section{Method}
\begin{figure}[t] 
    \centering
    \includegraphics[width=1.0\textwidth]{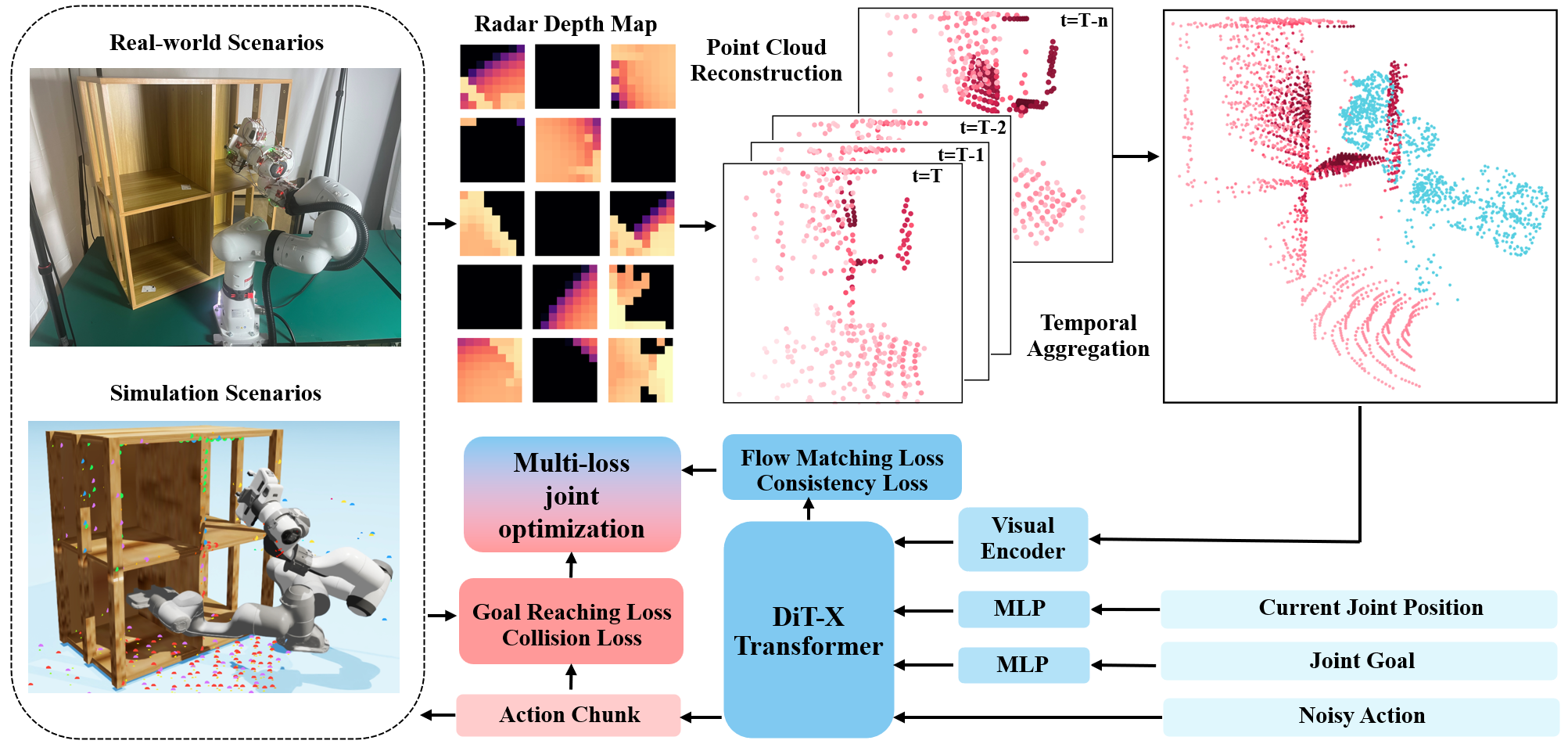} 
    \caption{ESCAPE is an online flow-matching policy for safe motion generation in cluttered environments. First, distributed body-mounted depth measurements are temporally aggregated into a structured, collision-aware point cloud. Then, a DiT-X Transformer takes this egocentric representation, the joint goal, and the current joint position as inputs to generate action chunks. The framework is trained via multi-loss joint optimization to output safe trajectories.}
    \label{fig:overall_pipeline}
\end{figure}
We propose ESCAPE, an online generative motion policy based on flow matching that produces stable, collision-free trajectories from distributed and sparse egocentric sensory data. An overview of our system architecture is illustrated in Figure~\ref{fig:overall_pipeline}.

From a manifold learning perspective, our framework learns a perception-conditioned vector field that maps a simple prior distribution onto a safe, obstacle-avoiding manifold. Unlike traditional Behavior Cloning, which often suffers from geometric rigidity in narrow spaces, ESCAPE provides physics-aware corrective gradients near obstacles. This mechanism enables trajectories to adaptively deform under complex geometric constraints, endowing the policy with strong generalization capabilities to robustly handle out-of-distribution scenarios in extremely cluttered environments.

To support policy learning from sparse body-mounted Time-of-Flight (ToF) sensing, we first convert distributed local measurements into a structured, collision-aware point cloud through temporal aggregation. The policy is trained on a large-scale dataset comprising over 100K collision-free trajectories, synthesized by coupling an obstacle-proximal endpoint sampling strategy with the cuRobo planner~\citep{sundaralingam2023curobo}. End-to-end learning is then performed with joint optimization of goal-reaching and collision objectives, followed by offline fine-tuning to improve robustness near obstacle boundaries.

\subsection{High-Quality Data Generation Protocol}
\label{sec:data_generation}

\begin{figure}[htbp] 
  \begin{minipage}{0.48\textwidth}
    \centering
    \includegraphics[width=\linewidth, trim=100 18 95 18, clip]{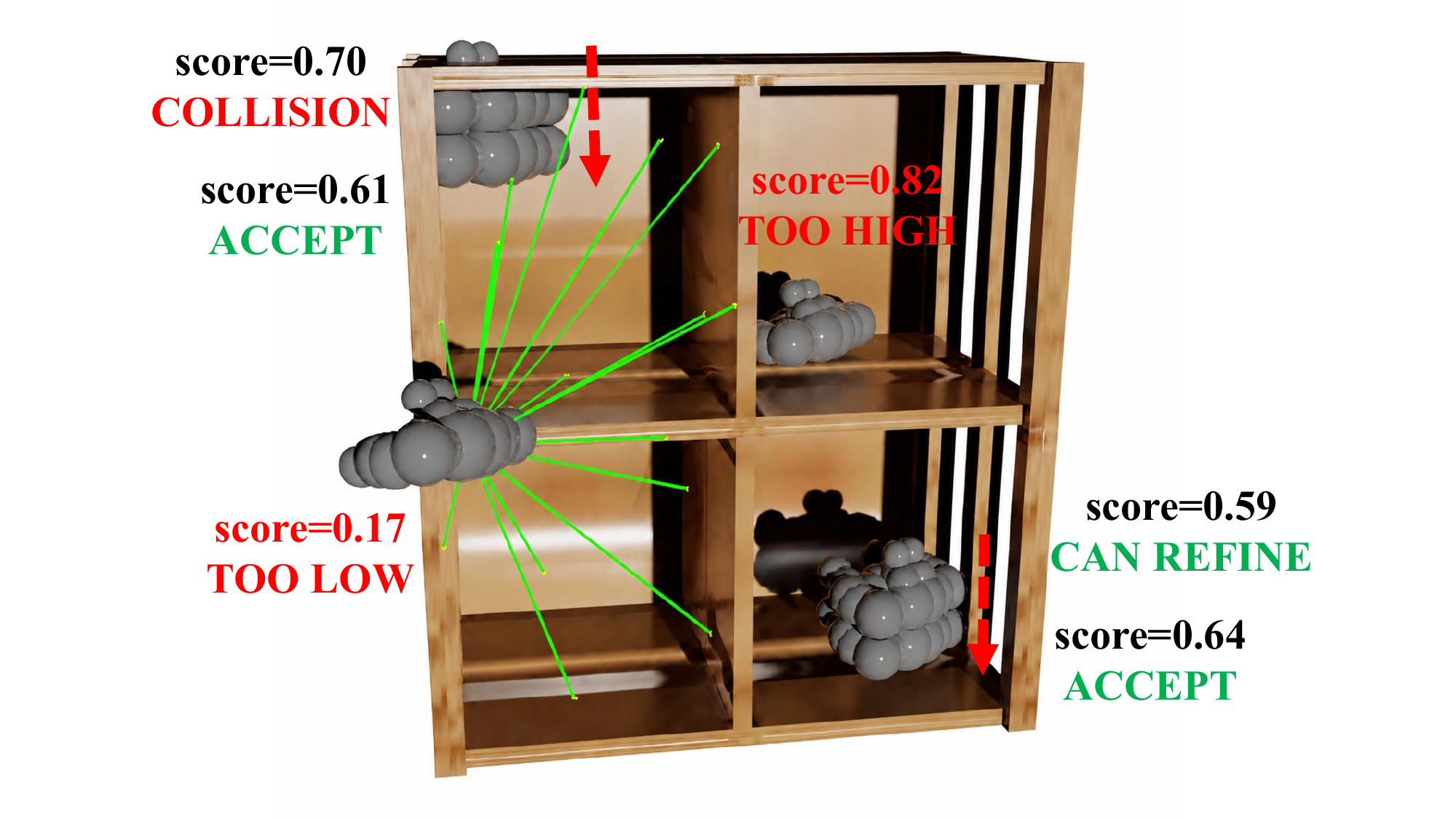}
    \caption{Obstacle-aware endpoint sampling. Candidate poses are evaluated via ray casting and filtered based on collision clearance.}
    \label{fig:sampling_protocol}
  \end{minipage}\hfill
  \begin{minipage}{0.48\textwidth}
    \centering
    \includegraphics[width=\linewidth]{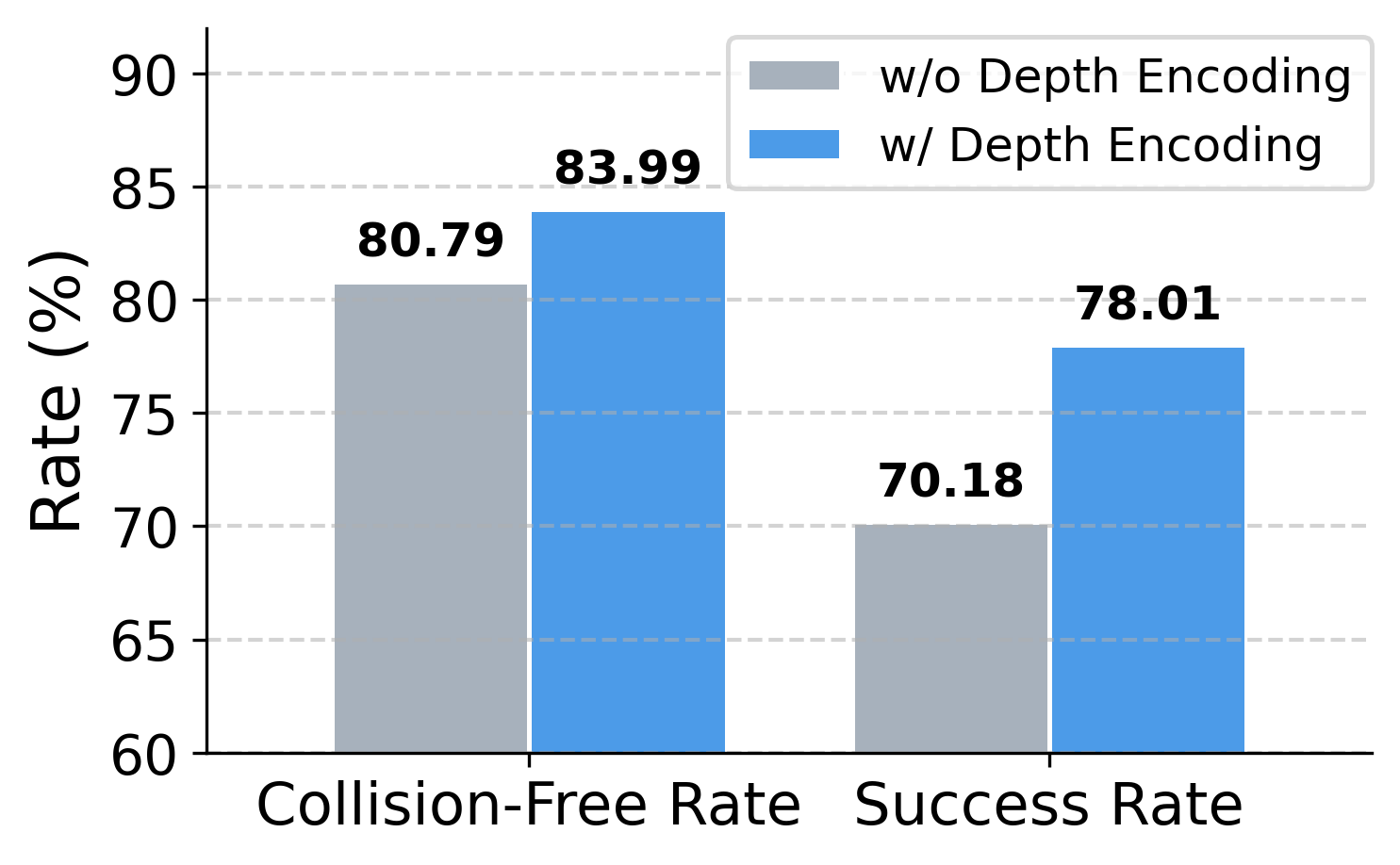}
    \caption{Depth encoding ablation. Proximity-aware features enhance near-field safety and overall task success.}
    \label{fig:ablation_depth}
  \end{minipage}
\end{figure}

To train a policy from egocentric perception, we generate collision-free demonstrations concentrated near safety-critical obstacle boundaries. 
The pipeline consists of obstacle-proximal endpoint sampling, continuous expert demonstration generation, synchronized egocentric observation collection, and offline corrective fine-tuning.
Figure~\ref{fig:sampling_protocol} illustrates the endpoint sampling procedure.

\textbf{Obstacle-proximal endpoint sampling.}
For controlled comparison with prior motion-planning benchmarks, we construct our simulation tasks from target-augmented DRP scenes~\citep{yang2025deep}. 
We sample candidate end-effector poses and score them by multi-directional ray casting, rejecting poses that are either too far from obstacles to be challenging or too close to maintain collision clearance. 
Refinement and overlap checks retain boundary-proximal, collision-free endpoints.

\textbf{Continuous demonstration and observation collection.}
For each validated endpoint, cuRobo~\citep{sundaralingam2023curobo} generates collision-aware joint-space trajectories. 
We retain trajectories satisfying success and distance constraints, chain them by using each terminal state as the next initial state, and replay them in simulation to record synchronized states, goals, expert actions, ToF point clouds, and raw body-mounted depth data. 
Applying this protocol to seven DRP-adapted scene categories with 20 demonstrations per instance yields over 100K collision-free expert trajectories with egocentric observations. 
Training on broader procedural environments remains future work.

\textbf{Offline corrective fine-tuning.}
After initial training, we perform DAgger-style corrective fine-tuning~\citep{ross2011reduction} to improve robustness near off-nominal states. 
We roll out the learned policy in simulation, collect collision-prone states, and use cuRobo~\citep{sundaralingam2023curobo} to generate corrective demonstrations from rollback configurations. 
These additional samples are aggregated into the training set to improve recovery near obstacle boundaries. 
Appendix~\ref{sec:rollback_ablation} studies different rollback strategies.

\subsection{Egocentric Perception Representation}
\label{sec:egocentric_perception}

ESCAPE utilizes a distributed array of ToF sensors integrated into the robot arm links for near-field observations. Due to the limited discrete points in single-frame readings, this distributed configuration faces the challenge of extreme spatial sparsity, making it difficult to directly characterize the complete geometric profile of obstacles.

To address this, we employ Proximity-aware Feature Encoding, which injects the raw relative depth values from the ToF sensors as an additional feature channel into the point cloud, guiding the model to focus on high-risk regions adjacent to the robot links and enhancing sensitivity to near-field collisions. As demonstrated in Figure \ref{fig:ablation_depth}, incorporating this depth encoding effectively boosts the collision-free rate and overall success rate. 

Subsequently, the local observations are projected into the global base frame via \(\mathbf{q}_t\) and forward kinematics, and self-filtering removes self-link points to obtain a clean environment point cloud. To increase perception density, we apply dilated temporal aggregation: a windowed aggregation with a progressive sampling stride that prioritizes the latest observations and increases the sampling interval with historical depth. This transforms time-distributed sparse signals into dense spatial contours, effectively profiling static obstacles while avoiding ``ghosting artifacts'' from ego-motion.

Finally, we perform structured resampling and semantic enhancement: we retain the closest environment points, and generate robot arm points via non-uniform sampling (dense near the end-effector, sparse near the base). We assign distinct semantic labels to both point sets via a color encoding channel. Both sets are transformed to the end-effector frame and concatenated as \(\mathcal{P}_{agg}\), providing the flow matching framework with explicit geometric boundaries and collision-aware constraints.

\subsection{Field-level Flow Matching}
\label{subsec:flow_matching}
Following ManiFlow~\cite{yan2025maniflow}, we model action generation as a flow ODE forward process. Given an expert action chunk $x_1 \sim \mathcal{D}$, a prior noise $x_0 \sim \mathcal{N}(0, \mathbf{I})$, and a timestep $t \sim \mathcal{U}[0,1]$, we define the intermediate state via linear interpolation $x_t = (1-t)x_0 + t x_1$. The target velocity field driving this path is $v_t = x_1 - x_0$. To unify velocity regression with trajectory self-consistency, we parameterize our policy network as $v_\theta(x_t, t, \Delta t, c)$, where $c$ denotes the conditioning context encompassing the egocentric perceptual representation, the current joint position, and the target goal. We optimize $v_\theta$ by jointly minimizing a conditional Flow Matching loss $\mathcal{L}_{fm}$ and a consistency loss $\mathcal{L}_{cons}$:
\begin{equation}
    \mathcal{L}_{fm}(\theta) = \mathbb{E}_{t, x_0, x_1} \left[ \| v_\theta(x_t, t, 0, c) - (x_1 - x_0) \|_2^2 \right]
\end{equation}
where setting $\Delta t = 0$ targets instantaneous velocity. To enforce self-consistency across temporal intervals, we concurrently minimize:
\begin{equation}
    \mathcal{L}_{cons}(\theta) = \mathbb{E}_{t, \Delta t, x_0, x_1} \left[ \| v_\theta(x_t, t, \Delta t, c) - \tilde{v}_{target} \|_2^2 \right]
\end{equation}
where $\tilde{v}_{target}$ is derived from an Exponential Moving Average (EMA) model. This field-level optimization circumvents iterative sequence generation, ensuring efficient training over massive datasets, while the consistency loss reduces the number of inference steps and accelerates inference.

\subsection{Trajectory-level Differentiable Unrolling}
\label{subsec:unrolling}

While field-level supervision fits the data distribution, it lacks explicit geometric awareness. To enforce strict safety, we introduce a differentiable unrolling mechanism. Starting from a sampled state \(x_t\) (\(t \sim \mathcal{U}[0,1]\)), we iteratively solve the flow ODE to predict the action chunk \(\hat{\tau} = [a_{t+1}, \dots, a_{t+H}]\) at \(x_1\). 
This differentiable forward pass enables task-level constraints, with step-count ablations detailed in Appendix~\ref{sec:ablation_steps}.
Specifically, to ensure continuous avoidance, we reconstruct a Signed Distance Field (SDF) using $\mathcal{P}_{agg}$. For each action $a_{t+j}$ in $\hat{\tau}$, forward kinematics projects the robot to 3D surface points $P(a_{t+j})$. Given safety margin $m$, the penalty is defined as:
\begin{figure}[t]
    \centering
    \includegraphics[width=1.0\textwidth]{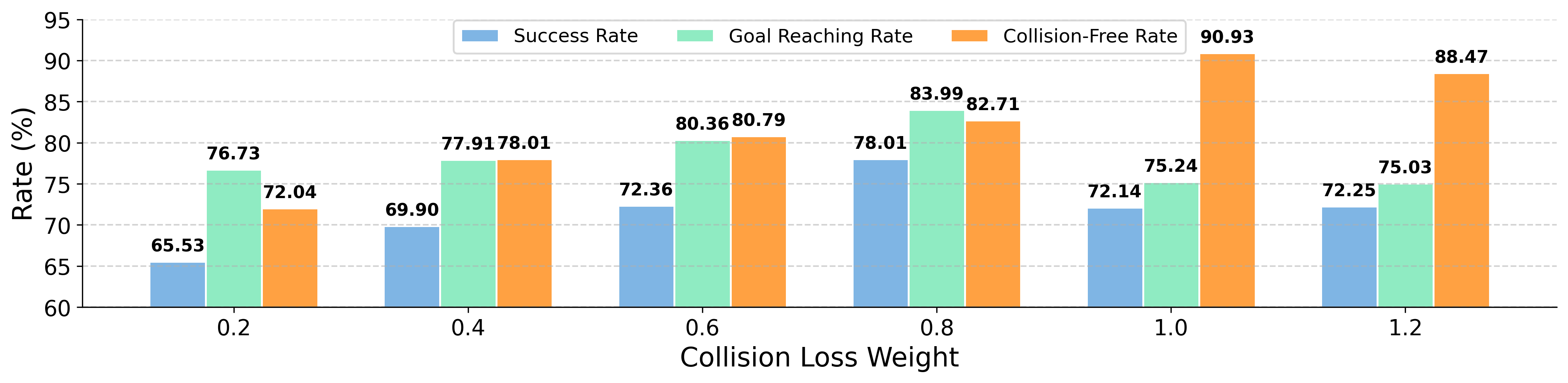}
    \caption{Ablation of collision loss weight. As the weight increases, the collision-free rate initially rises before eventually declining, with a weight of 0.8 achieving the best overall performance.}
    \label{fig:ablation_collision}
\end{figure}

\begin{equation}
    \mathcal{L}_{coll} = \frac{1}{H \cdot N} \sum_{j=1}^{H} \sum_{k=1}^{N} \left[ \frac{\max(0, m - \mathrm{SDF}(p_{j,k}))}{m} \right]^2
\end{equation}
where $H$ denotes the action chunk horizon, $N$ is the number of sampled robot surface points, and $p_{j,k}$ represents the coordinate of the $k$-th surface point at time step $j$. The physical repulsive gradients computed from this loss backpropagate through the solver graph to shape the vector field, forcing trajectories to adaptively deform around obstacle boundaries. As shown in Fig.~\ref{fig:ablation_collision}, increasing the collision loss weight initially improves both goal-reaching and collision-free rates, because severe collisions often stall the robot and hurt goal attainment. 
However, excessively high weights cause over-conservative avoidance and gradient convergence issues, ultimately degrading the terminal goal-reaching and total success rates even as safety improves. Consequently, we select a collision loss weight of 0.8 to achieve an optimal balance between trajectory safety and goal convergence.

\begin{figure}[t]
    \centering
    \includegraphics[width=1.0\textwidth]{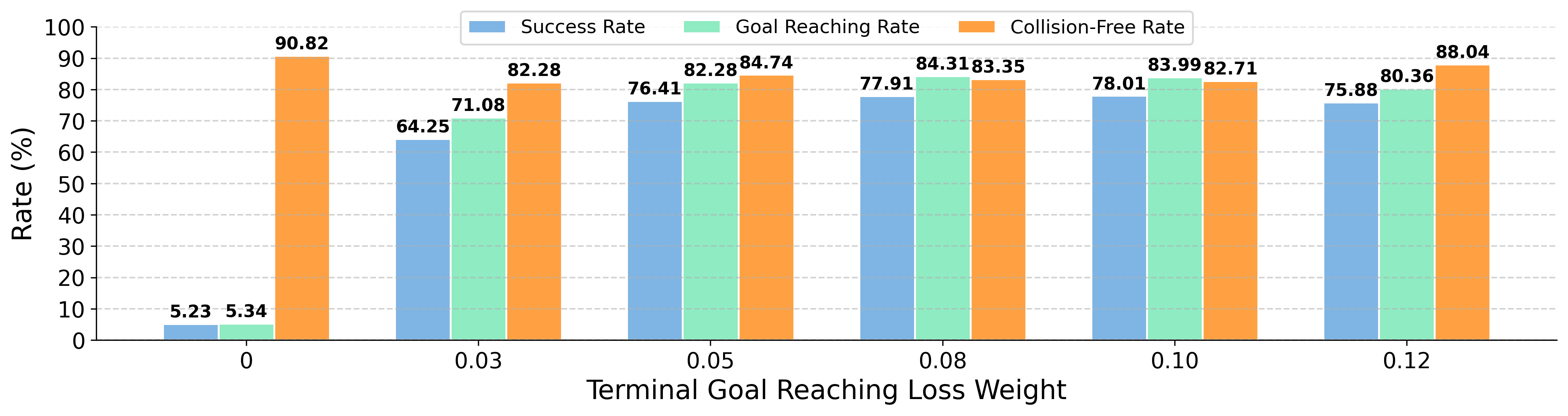}
    \caption{Ablation of terminal goal-reaching loss weight. As the weight increases, the goal-reaching rate initially rises significantly before slowly declining, with 0.1 achieving the highest success rate.}
    \label{fig:ablation_goal}
\end{figure}

Further, to balance the rigid spatial constraints of collision avoidance with the gradually evolving nature of goal reaching, we apply a sparse goal-reaching loss that activates only within a terminal window of \(K=15\) steps before episode end. For a sample with \(r\) remaining steps, the loss penalizes the last \(t_{\text{tail}} = \min(H, K - r + 1)\) steps of the predicted action chunk (where \(H=15\)), so that more steps are penalized as the robot approaches the goal:
\begin{equation}
    \mathcal{L}_{\text{goal}} = \sum_{j = H - t_{\text{tail}} + 1}^{H} \left\| \hat{\mathbf{q}}_{t+j} - \mathbf{q}_{\text{goal}} \right\|_1
\end{equation}
where the loss computes an L1 penalty in the original joint space between the predicted joint states \(\hat{\mathbf{q}}_{t+j}\) and the target \(\mathbf{q}_{\text{goal}}\). This design ensures that the policy balances collision avoidance gradients early in the rollout with asymptotic convergence to the target. Our reproducibility analysis in Fig.~\ref{fig:ablation_goal} highlights this systemic relationship: while a baseline goal-reaching loss is required to guide the end-effector precisely to the target, excessively scaling this objective creates a gradient conflict with collision avoidance, disrupting the trajectory deformation near safety boundaries. Consequently, stable motion generation relies on a joint optimization of $\mathcal{L}_{coll}$ and $\mathcal{L}_{goal}$ that respects the balance between safety constraints and goal attraction. Importantly, we can safely and efficiently map out this optimal setting purely within simulation without demanding real-world parameter tuning.


\section{Experimental Results}
\label{sec:result}

\subsection{Simulation Experiments and Results}
\label{subsec:simulation_results}

We evaluate ESCAPE on a unified simulation benchmark of 937 Franka reaching problems in static cluttered environments. 
Each problem specifies an initial robot configuration and a target end-effector pose, and success requires reaching the target within position and orientation thresholds while remaining collision-free.

We compare ESCAPE with privileged planners, global-observation policies, and partial-observation policies on our benchmark. 
AIT* and cuRobo use privileged scene information. 
M$\pi$Nets, M$\pi$Former, NeuralMP, cuRobo-Vox, and DRP-G use global or map-based scene observations, while DRP-Cam and ESCAPE operate from partial observations. 
For scene perception, ESCAPE uses only robot-mounted near-field sensing.
\begin{table}[!htbp]
  \centering
  \caption{Success rate comparison on the unified simulation benchmark.}
  \label{tab:simulation_comparison}
  \scriptsize
  \setlength{\tabcolsep}{2.8pt}
  \renewcommand{\arraystretch}{1.16}
  \resizebox{0.97\textwidth}{!}{%
  \begin{tabular}{@{}lcccccccccc@{}}
    \toprule
    &
    \multicolumn{2}{c}{Privileged Planning} &
    \multicolumn{5}{c}{Global / Map Observation} &
    \multicolumn{3}{c}{Partial / Egocentric Observation} \\
    \cmidrule(lr){2-3}
    \cmidrule(lr){4-8}
    \cmidrule(lr){9-11}
    Test Split
    & AIT*
    & cuRobo
    & M$\pi$Nets
    & M$\pi$F-T
    & NeuralMP
    & \shortstack{cuRobo\\Vox}
    & DRP-G
    & DRP-Cam
    & \shortstack{ESCAPE\\w/o FT}
    & ESCAPE \\
    \midrule
    Easy
    & 74.95\%
    & \textbf{95.76\%}
    & 1.01\%
    & 13.74\%
    & 55.96\%
    & 82.83\%
    & 85.45\%
    & 66.06\%
    & 90.70\%
    & 90.30\% \\

    Hard
    & 11.99\%
    & 55.66\%
    & 4.75\%
    & 10.18\%
    & 41.18\%
    & 26.92\%
    & 72.40\%
    & 45.93\%
    & 63.80\%
    & \textbf{76.24\%} \\

    Total
    & 45.25\%
    & 76.84\%
    & 2.77\%
    & 12.06\%
    & 48.99\%
    & 56.46\%
    & 79.30\%
    & 56.56\%
    & 78.02\%
    & \textbf{83.67\%} \\
    \bottomrule
  \end{tabular}%
  }

  \footnotesize{
  M$\pi$F-T denotes M$\pi$Former-Tabletop.
  cuRobo-Vox~\citep{sundaralingam2023curobo} denotes cuRobo with a voxelized collision representation.
  DRP-G denotes DRP with global point clouds, and DRP-Cam denotes DRP with iterative external-camera observations.
  ESCAPE w/o FT denotes ESCAPE before offline fine-tuning.
  }
\end{table}

\textbf{ESCAPE achieves the strongest overall simulation performance.}
As shown in Table~\ref{tab:simulation_comparison}, ESCAPE obtains the best Hard and Total success rates, reaching 76.24\% and 83.67\%, respectively. 
It remains close to privileged cuRobo on Easy tasks, while using only robot-mounted egocentric sensing rather than privileged geometry or global scene observations. 
Compared with DRP-G and DRP-Cam, ESCAPE improves the overall success rate by 4.37 and 27.11 percentage points, respectively.

\textbf{Offline fine-tuning improves hard-case robustness.}
Offline fine-tuning increases the Hard success rate from 63.80\% to 76.24\% and the overall success rate from 78.02\% to 83.67\%. 
This gain suggests that rollback data helps the policy recover from off-nominal states near obstacle boundaries and narrow feasible regions.


\subsection{Real-World Experiments and Results}
\label{subsec:real_world_results}
We evaluate ESCAPE on a real Franka robot across six cluttered reaching scenes: ShelfEasy, ShelfHard, WallCabinet, Box, Cage, and Microwave (See Fig.~\ref{fig:figure1}), with 92 valid trials in total. The number of trials varies across these environments to appropriately match the specific geometric complexity of each scene.
A trial is successful if the robot reaches the target without task-disrupting contact, protective stop, execution interruption, or visible object displacement. 
We compare ESCAPE with NeuralMP, cuRobo-Vox, and DRP under the same task set, where ESCAPE is trained entirely in simulation and deployed zero-shot using only robot-mounted near-field sensing.

\textbf{ESCAPE achieves strong zero-shot real-world transfer.}
As shown in Table~\ref{tab:real_world_comparison}, ESCAPE succeeds in 84 out of 92 trials, achieving the highest overall success rate of 91.30\% and the best  result in every scene. 
It outperforms DRP and cuRobo-Vox by 19.56 and 43.47 percentage points, respectively.

\begin{table}[t]
  \centering
  \caption{Zero-shot real-world success rate comparison on the six-scene evaluation set.}
  \label{tab:real_world_comparison}
  \small
  \renewcommand{\arraystretch}{1.15}
  \begin{tabularx}{0.97\textwidth}{@{}l *{6}{>{\centering\arraybackslash}X} >{\centering\arraybackslash}p{0.15\textwidth}@{}}
    \toprule
    Method & Shelf-E & Shelf-H & WallCab. & Box & Cage & Micro. & Total \\
    \midrule
    NeuralMP
    & 0/15
    & 2/18
    & 0/18
    & 5/18
    & 0/13
    & 1/10
    & \mbox{8/92 (8.70\%)} \\

    cuRobo-Vox
    & 9/15
    & 8/18
    & 10/18
    & 3/18
    & 10/13
    & 4/10
    & \mbox{44/92 (47.83\%)} \\

    DRP
    & \textbf{13/15}
    & 12/18
    & 11/18
    & 11/18
    & 12/13
    & 7/10
    & \mbox{66/92 (71.74\%)} \\

    ESCAPE
    & \textbf{13/15}
    & \textbf{15/18}
    & \textbf{17/18}
    & \textbf{18/18}
    & \textbf{13/13}
    & \textbf{8/10}
    & \mbox{\textbf{84/92 (91.30\%)}} \\
    \bottomrule
  \end{tabularx}
\end{table}

\textbf{Robot-mounted sensing simplifies deployment.}
Unlike external-camera methods, ESCAPE does not require careful camera placement, multi-camera calibration, or scene-dependent view adjustment, and is less vulnerable to accidental camera displacement during deployment.
The robot-mounted egocentric sensing setup keeps perception tied to the robot itself, enabling a plug-and-play sim-to-real deployment once the policy is trained in simulation.


\section{Conclusion}
\label{sec:conclusion}
We introduce ESCAPE, an online egocentric motion-generation framework for collision-free reaching in cluttered and partially observed environments. 
By mounting distributed depth sensors directly on the robot, ESCAPE reduces reliance on external-camera placement, calibration, and view stability. 
A temporal aggregation mechanism converts sparse body-mounted ToF measurements into an obstacle-avoidance-aware point cloud, which conditions a flow matching-based generative policy trained entirely in simulation. 
With joint optimization of goal-reaching and collision objectives, the policy generates safe, goal-directed trajectories from egocentric observations, robot state, and goals. 
Without real-world fine-tuning, ESCAPE achieves strong zero-shot performance on a Franka robot and outperforms external-camera and voxelized-planning baselines across cluttered reaching tasks. 
These results suggest that robot-mounted egocentric sensing provides a practical path toward plug-and-play collision-aware manipulation without carefully engineered external perception setups.
\section{Limitations}
ESCAPE presents several areas for improvement. 
First, our current evaluation focuses on static cluttered reaching tasks. 
The temporal aggregation module is designed to densify sparse egocentric observations of static near-field geometry; tackling moving obstacles is left as a future direction. 
Second, the distributed ToF sensors provide sparse depth measurements and can be affected by challenging surface properties, such as highly reflective, transparent, or dark materials. Including other sensing modalities would expand ESCAPE's robustness to diverse material semantics.
Third, the current hardware design and policy are evaluated on a single robot embodiment, the Franka arm; transferring the system to other platforms would require redesigned sensor mounts, recalibrated sensor frames, and additional training. 
Future work will extend the environment distribution beyond the current set of scenes, and advance ESCAPE toward dynamic human-robot collaboration spaces.

\clearpage


\bibliography{example}  

@inproceedings{strub2020adaptively,
  title={Adaptively Informed Trees (AIT*): Fast asymptotically optimal path planning through adaptive heuristics},
  author={Strub, Marlin P and Gammell, Jonathan D},
  booktitle={2020 IEEE International Conference on Robotics and Automation (ICRA)},
  pages={3191--3198},
  year={2020},
  organization={IEEE}
}

@ARTICLE{4082128,
  author={Hart, Peter E. and Nilsson, Nils J. and Raphael, Bertram},
  journal={IEEE Transactions on Systems Science and Cybernetics}, 
  title={A Formal Basis for the Heuristic Determination of Minimum Cost Paths}, 
  year={1968},
  volume={4},
  number={2},
  pages={100-107},
  doi={10.1109/TSSC.1968.300136}}

@article{karaman2011sampling,
  title={Sampling-based algorithms for optimal motion planning},
  author={Karaman, Sertac and Frazzoli, Emilio},
  journal={The international journal of robotics research},
  volume={30},
  number={7},
  pages={846--894},
  year={2011},
  publisher={Sage Publications Sage UK: London, England}
}

@article{orthey2023sampling,
  title={Sampling-based Motion Planning: A Comparative Review},
  author={Orthey, Andreas and Chamzas, Constantinos and Kavraki, Lydia E.},
  journal={Annual Review of Control, Robotics, and Autonomous Systems},
  volume={7},
  number={1},
  pages={285--310},
  year={2024},
  publisher={Annual Reviews},
  doi={10.1146/annurev-control-061623-094742}
}

@article{sundaralingam2023curobo,
  title={curobo: Parallelized collision-free minimum-jerk robot motion generation},
  author={Sundaralingam, Balakumar and Hari, Siva Kumar Sastry and Fishman, Adam and Garrett, Caelan and Van Wyk, Karl and Blukis, Valts and Millane, Alexander and Oleynikova, Helen and Handa, Ankur and Ramos, Fabio and others},
  journal={arXiv preprint arXiv:2310.17274},
  year={2023}
}

@inproceedings{fishman2023motion,
  title={Motion policy networks},
  author={Fishman, Adam and Murali, Adithyavairavan and Eppner, Clemens and Peele, Bryan and Boots, Byron and Fox, Dieter},
  booktitle={conference on Robot Learning},
  pages={967--977},
  year={2023},
  organization={PMLR}
}

@InProceedings{pmlr-v270-fishman25a,
  title = 	 {Avoid Everything: Model-Free Collision Avoidance with Expert-Guided Fine-Tuning},
  author =       {Fishman, Adam and Walsman, Aaron and Bhardwaj, Mohak and Yuan, Wentao and Sundaralingam, Balakumar and Boots, Byron and Fox, Dieter},
  booktitle = 	 {Proceedings of The 8th Conference on Robot Learning},
  pages = 	 {1925--1948},
  year = 	 {2025},
  editor = 	 {Agrawal, Pulkit and Kroemer, Oliver and Burgard, Wolfram},
  volume = 	 {270},
  series = 	 {Proceedings of Machine Learning Research},
  month = 	 {06--09 Nov},
  publisher =    {PMLR},
  url = 	 {https://proceedings.mlr.press/v270/fishman25a.html}
}

@article{dalal2024neural,
  title={Neural mp: A generalist neural motion planner},
  author={Dalal, Murtaza and Yang, Jiahui and Mendonca, Russell and Khaky, Youssef and Salakhutdinov, Ruslan and Pathak, Deepak},
  journal={arXiv preprint arXiv:2409.05864},
  year={2024}
}

@article{yang2025deep,
  title={Deep reactive policy: Learning reactive manipulator motion planning for dynamic environments},
  author={Yang, Jiahui and Liu, Jason Jingzhou and Li, Yulong and Khaky, Youssef and Shaw, Kenneth and Pathak, Deepak},
  journal={arXiv preprint arXiv:2509.06953},
  year={2025}
}

@article{kim2024armor,
  title={Armor: Egocentric perception for humanoid robot collision avoidance and motion planning},
  author={Kim, Daehwa and Srouji, Mario and Chen, Chen and Zhang, Jian},
  journal={arXiv preprint arXiv:2412.00396},
  year={2024}
}

@article{singletary2022safety,
  title={Safety-critical manipulation for collision-free food preparation},
  author={Singletary, Andrew and Guffey, William and Molnar, Tamas G and Sinnet, Ryan and Ames, Aaron D},
  journal={IEEE Robotics and Automation Letters},
  volume={7},
  number={4},
  pages={10954--10961},
  year={2022},
  publisher={IEEE}
}

@inproceedings{thirugnanam2022safety,
  title={Safety-critical control and planning for obstacle avoidance between polytopes with control barrier functions},
  author={Thirugnanam, Akshay and Zeng, Jun and Sreenath, Koushil},
  booktitle={2022 International Conference on Robotics and Automation (ICRA)},
  pages={286--292},
  year={2022},
  organization={IEEE}
}

@inproceedings{yu2024efficient,
  title={Efficient motion planning for manipulators with control barrier function-induced neural controller},
  author={Yu, Mingxin and Yu, Chenning and Naddaf-Sh, M-Mahdi and Upadhyay, Devesh and Gao, Sicun and Fan, Chuchu},
  booktitle={2024 IEEE international conference on robotics and automation (ICRA)},
  pages={14348--14355},
  year={2024},
  organization={IEEE}
}

@article{likhachev2003ara,
  title={ARA*: Anytime A* with provable bounds on sub-optimality},
  author={Likhachev, Maxim and Gordon, Geoffrey J and Thrun, Sebastian},
  journal={Advances in neural information processing systems},
  volume={16},
  year={2003}
}

@inproceedings{likhachev2005anytime,
  title={Anytime dynamic A*: An anytime, replanning algorithm.},
  author={Likhachev, Maxim and Ferguson, David I and Gordon, Geoffrey J and Stentz, Anthony and Thrun, Sebastian},
  booktitle={ICAPS},
  volume={5},
  pages={262--271},
  year={2005}
}

@inproceedings{koenig2006new,
  title={A New Principle for Incremental Heuristic Search: Theoretical Results.},
  author={Koenig, Sven and Likhachev, Maxim},
  booktitle={ICAPS},
  pages={402--405},
  year={2006}
}

@article{kavraki2002probabilistic,
  title={Probabilistic Roadmaps for Path Planning in High-Dimensional Configuration Spaces},
  author={Kavraki, Lydia E. and Svestka, Petr and Latombe, Jean-Claude and Overmars, Mark H.},
  journal={IEEE Transactions on Robotics and Automation},
  volume={12},
  number={4},
  pages={566--580},
  year={1996},
  publisher={IEEE},
  doi={10.1109/70.508439}
}

@techreport{trees1998new,
  title={Rapidly-Exploring Random Trees: A New Tool for Path Planning},
  author={LaValle, Steven M.},
  institution={Department of Computer Science, Iowa State University},
  number={TR 98-11},
  year={1998},
  url={https://msl.cs.illinois.edu/~lavalle/papers/Lav98c.pdf}
}

@inproceedings{ross2011reduction,
  title={A Reduction of Imitation Learning and Structured Prediction to No-Regret Online Learning},
  author={Ross, Stephane and Gordon, Geoffrey and Bagnell, Drew},
  booktitle={Proceedings of the Fourteenth International Conference on Artificial Intelligence and Statistics},
  pages={627--635},
  year={2011},
  volume={15},
  series={Proceedings of Machine Learning Research},
  publisher={PMLR}
}

@article{zucker2013chomp,
  title={Chomp: Covariant hamiltonian optimization for motion planning},
  author={Zucker, Matt and Ratliff, Nathan and Dragan, Anca D and Pivtoraiko, Mihail and Klingensmith, Matthew and Dellin, Christopher M and Bagnell, J Andrew and Srinivasa, Siddhartha S},
  journal={The International journal of robotics research},
  volume={32},
  number={9-10},
  pages={1164--1193},
  year={2013},
  publisher={SAGE Publications Sage UK: London, England}
}

@inproceedings{dragan2011manipulation,
  title={Manipulation planning with goal sets using constrained trajectory optimization},
  author={Dragan, Anca D and Ratliff, Nathan D and Srinivasa, Siddhartha S},
  booktitle={2011 IEEE International Conference on Robotics and Automation},
  pages={4582--4588},
  year={2011},
  organization={IEEE}
}

@article{schulman2014motion,
  title={Motion planning with sequential convex optimization and convex collision checking},
  author={Schulman, John and Duan, Yan and Ho, Jonathan and Lee, Alex and Awwal, Ibrahim and Bradlow, Henry and Pan, Jia and Patil, Sachin and Goldberg, Ken and Abbeel, Pieter},
  journal={The International Journal of Robotics Research},
  volume={33},
  number={9},
  pages={1251--1270},
  year={2014},
  publisher={Sage Publications Sage UK: London, England}
}

@inproceedings{qureshi2019motion,
  title={Motion planning networks},
  author={Qureshi, Ahmed H and Simeonov, Anthony and Bency, Mayur J and Yip, Michael C},
  booktitle={2019 International Conference on Robotics and Automation (ICRA)},
  pages={2118--2124},
  year={2019},
  organization={IEEE}
}

@inproceedings{kumar2019lego,
  title={Lego: Leveraging experience in roadmap generation for sampling-based planning},
  author={Kumar, Rahul and Mandalika, Aditya and Choudhury, Sanjiban and Srinivasa, Siddhartha},
  booktitle={2019 IEEE/RSJ International Conference on Intelligent Robots and Systems (IROS)},
  pages={1488--1495},
  year={2019},
  organization={IEEE}
}

@inproceedings{zhang2018learning,
  title={Learning implicit sampling distributions for motion planning},
  author={Zhang, Clark and Huh, Jinwook and Lee, Daniel D},
  booktitle={2018 IEEE/RSJ International Conference on Intelligent Robots and Systems (IROS)},
  pages={3654--3661},
  year={2018},
  organization={IEEE}
}

@inproceedings{chamzas2021learning,
  title={Learning sampling distributions using local 3d workspace decompositions for motion planning in high dimensions},
  author={Chamzas, Constantinos and Kingston, Zachary and Quintero-Pe{\~n}a, Carlos and Shrivastava, Anshumali and Kavraki, Lydia E},
  booktitle={2021 IEEE International Conference on Robotics and Automation (ICRA)},
  pages={1283--1289},
  year={2021},
  organization={IEEE}
}

@article{ichter2020broadly,
  title={Broadly-exploring, local-policy trees for long-horizon task planning},
  author={Ichter, Brian and Sermanet, Pierre and Lynch, Corey},
  journal={arXiv preprint arXiv:2010.06491},
  year={2020}
}

@article{qureshi2020neural,
  title={Neural manipulation planning on constraint manifolds},
  author={Qureshi, Ahmed H and Dong, Jiangeng and Choe, Austin and Yip, Michael C},
  journal={IEEE Robotics and Automation Letters},
  volume={5},
  number={4},
  pages={6089--6096},
  year={2020},
  publisher={IEEE}
}

@article{xu2025robopanoptes,
  title={Robopanoptes: The all-seeing robot with whole-body dexterity},
  author={Xu, Xiaomeng and Bauer, Dominik and Song, Shuran},
  journal={arXiv preprint arXiv:2501.05420},
  year={2025}
}

@article{ding2025imagination,
  title={Imagination at Inference: Synthesizing In-Hand Views for Robust Visuomotor Policy Inference},
  author={Ding, Haoran and Duan, Anqing and Sun, Zezhou and Song, Dezhen and Nakamura, Yoshihiko},
  journal={arXiv preprint arXiv:2509.15717},
  year={2025}
}

@article{zou2026activeglasses,
  title={ActiveGlasses: Learning Manipulation with Active Vision from Ego-centric Human Demonstration},
  author={Zou, Yanwen and Shi, Chenyang and Yu, Wenye and Xue, Han and Lv, Jun and Pan, Ye and Wen, Chuan and Lu, Cewu},
  journal={arXiv preprint arXiv:2604.08534},
  year={2026}
}

@inproceedings{schlegl2013virtual,
  title={Virtual whiskers—Highly responsive robot collision avoidance},
  author={Schlegl, Thomas and Kr{\"o}ger, Torsten and Gaschler, Andre and Khatib, Oussama and Zangl, Hubert},
  booktitle={2013 IEEE/RSJ International Conference on Intelligent Robots and Systems},
  pages={5373--5379},
  year={2013},
  organization={IEEE}
}

@article{tsuji2019proximity,
  title={Proximity skin sensor using time-of-flight sensor for human collaborative robot},
  author={Tsuji, Satoshi and Kohama, Teruhiko},
  journal={IEEE Sensors Journal},
  volume={19},
  number={14},
  pages={5859--5864},
  year={2019},
  publisher={IEEE}
}

@article{giovinazzo2024cyskin,
  title={From CySkin to ProxySKIN: Design, implementation and testing of a multi-modal robotic skin for human--robot interaction},
  author={Giovinazzo, Francesco and Grella, Francesco and Sartore, Marco and Adami, Manuela and Galletti, Riccardo and Cannata, Giorgio},
  journal={Sensors},
  volume={24},
  number={4},
  pages={1334},
  year={2024},
  publisher={MDPI}
}

@article{yan2025maniflow,
  title={Maniflow: A general robot manipulation policy via consistency flow training},
  author={Yan, Ge and Zhu, Jiyue and Deng, Yuquan and Yang, Shiqi and Qiu, Ri-Zhao and Cheng, Xuxin and Memmel, Marius and Krishna, Ranjay and Goyal, Ankit and Wang, Xiaolong and others},
  journal={arXiv preprint arXiv:2509.01819},
  year={2025}
}

\appendix
\section{Distributed Depth-Sensing Hardware Design}
\label{subsec:depth_sensing_hardware}

\begin{figure}[!htbp]
    \centering
    \includegraphics[width=\linewidth]{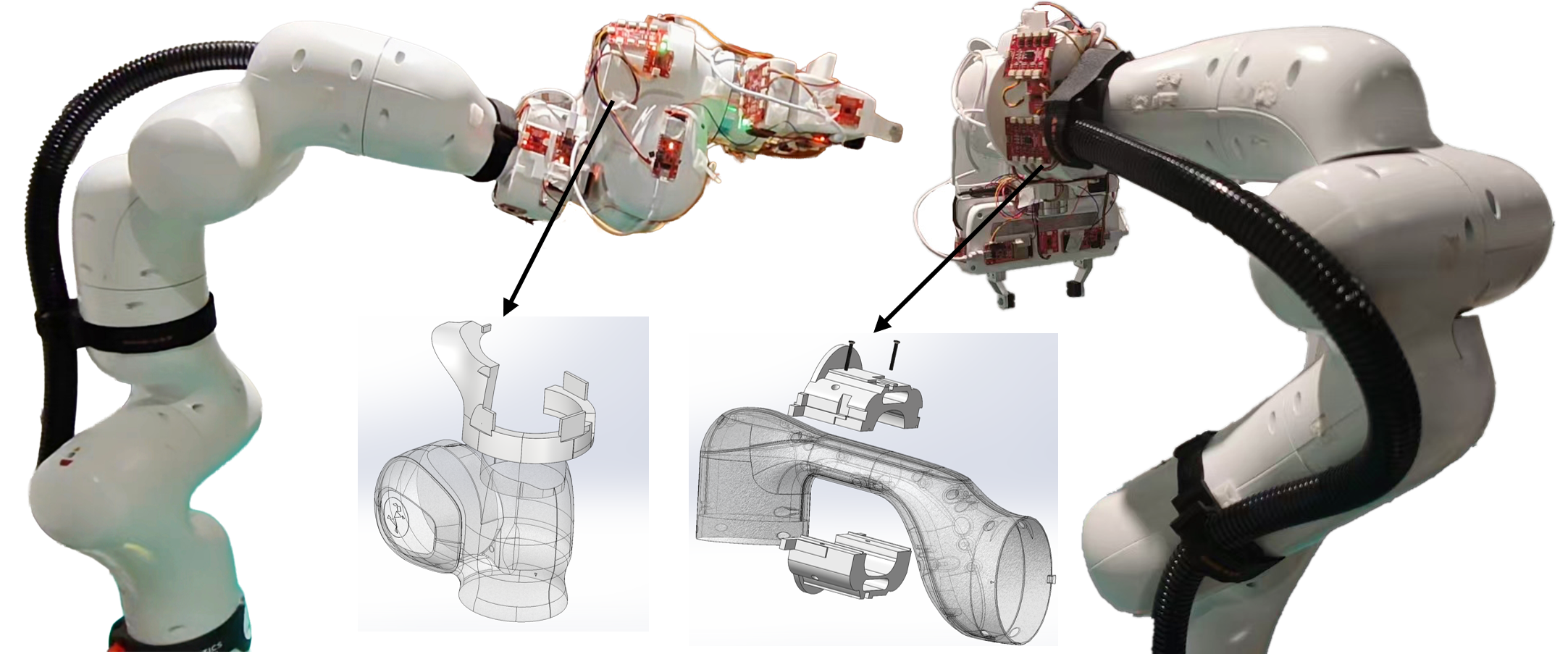}
    \caption{Distributed sensor mounting and integration on the Franka arm. The system features 3D-printed conformal brackets: a split-shell design for Link 5 (right) and a half-ring design for Link 6 (left). The real-world deployment showcases the wiring and integration of the ToF modules around the end-effector region.}
    \label{fig:hardware_overview}
\end{figure}
To support online multi-view depth perception for collision-aware motion generation, we deploy 15 VL53L5CX multi-zone ToF modules ($8\times8$ zones, $63^\circ$ FoV, 15\,Hz) on the Franka arm (Fig.~\ref{fig:hardware_overview}). Each group of three modules is aggregated by an ESP32-WROOM controller through a TCA9548A I\textsuperscript{2}C multiplexer for bus isolation. Five such units form the complete sensing network.

\textbf{Sensor mounting brackets.}
We design dedicated brackets for Link\,5 and Link\,6, with inner surfaces offset from the link geometry for a conformal fit. The Link\,5 bracket splits into two halves with mortise-and-tenon alignment and screw fastening; outward protrusions reduce self-occlusion, and sensor recesses improve pose repeatability for sim-to-real alignment. The Link\,6 bracket uses a half-ring nesting design fixed by friction alone. Together, the two brackets provide approximately $360^\circ$ local coverage around the end-effector region.

\textbf{Digital twin and physical deployment.}
The assembly is exported as a URDF model with calibrated local frames, inertia tensors, and RPY orientations, and is verified through TF-tree, kinematic-frame, and collision-mesh checks. As depicted in the real-world setup, the brackets are 3D-printed and integrated onto the Franka wrist, with the I\textsuperscript{2}C communication and power harnesses explicitly routed along the arm to maintain unobstructed kinematic mobility. System-level dynamic tests confirm both the structural stability of the sensor array and its pose repeatability. By redesigning the mounts and recalibrating the sensor frames, this wrist-mounted design can be readily adapted to other robot embodiments, including humanoid robots.

\section{Detailed Task Descriptions}
\label{app:task_descriptions}

We evaluate ESCAPE on a unified simulation benchmark of Franka reaching tasks in static cluttered environments. 
Each task is defined by a scene, an initial robot joint configuration, and a target end-effector pose. 
The robot must move from the initial configuration to the target pose while avoiding both environment collisions and self-collisions. 
All methods are evaluated on the same set of scenes, initial states, target poses, and success criteria.

\textbf{Scene categories.}
The benchmark is constructed from seven categories of static obstacle scenes adapted from the DRP environments, including shelf-like, cubby-like, wall-cabinet, microwave, dishwasher, box, and cage scenes. 
These scenes contain cluttered structures, narrow passages, and partially enclosed regions that require obstacle-aware motion generation. 
Unlike dynamic-obstacle benchmarks, our evaluation focuses on static but geometrically constrained reaching, where the key challenge is generating safe motions from partial or egocentric observations.

\textbf{Easy and Hard splits.}
We divide the benchmark into Easy and Hard subsets according to the target sampling difficulty. 
Easy tasks are sampled with relaxed obstacle-proximity constraints and skip the boundary-refinement stage; their target poses are typically located in more open regions with larger free space around the end effector. 
Hard tasks use the boundary-aware target sampling procedure described in Section~\ref{sec:data_generation}: candidate targets are filtered by obstacle proximity and further refined toward obstacle boundaries or narrow feasible regions. 
All accepted targets are collision-free, but Hard tasks require more precise local avoidance and closed-loop correction near cluttered geometry.

\textbf{Benchmark size.}
The final benchmark contains 937 task instances, including 495 Easy tasks and 442 Hard tasks. 
The Total score reported in the main paper is computed over all 937 tasks rather than by averaging the two split-level success rates. 
We avoid overlap between training and evaluation by using held-out task instances and by preventing highly similar trajectory fragments from appearing in both training and testing.

\textbf{Success metric.}
We follow the standard goal-reaching metric used in prior motion-planning benchmarks: a trial is counted as successful if the robot reaches the end-effector goal pose within a $1$ cm translational error and a $15^\circ$ orientation error, while remaining collision-free throughout execution. 
A trial is counted as failed if the robot collides with the environment or itself, fails to reach the goal threshold, or exceeds the rollout horizon. 

To ensure rigorous evaluation, we explicitly define the collision detection mechanism. For the baseline methods, collision status is determined by querying the environment's Signed Distance Field (SDF), where a collision is registered if $\text{SDF} \le 0$. In contrast, our ESCAPE framework fully enables physical collision entities within the simulation. Consequently, a collision is natively determined by the physical execution outcome: if the inferred joint command for the next time step cannot be successfully reached due to physical impedance from the environment, it is recorded as a collision. 

In addition to the main success rate, we record goal-reaching rate and collision-free rate in ablation studies to separate terminal accuracy from safety. Specifically, the goal-reaching rate accounts for cases where the robot, despite experiencing a physical collision that severely alters its trajectory, still manages to adjust and successfully reach the target pose. 
We also inspect stricter orientation thresholds as diagnostic metrics, while the main comparison follows the $15^\circ$ threshold for consistency with prior work.

\begin{figure}[t]
    \centering
    \includegraphics[width=0.95\linewidth]{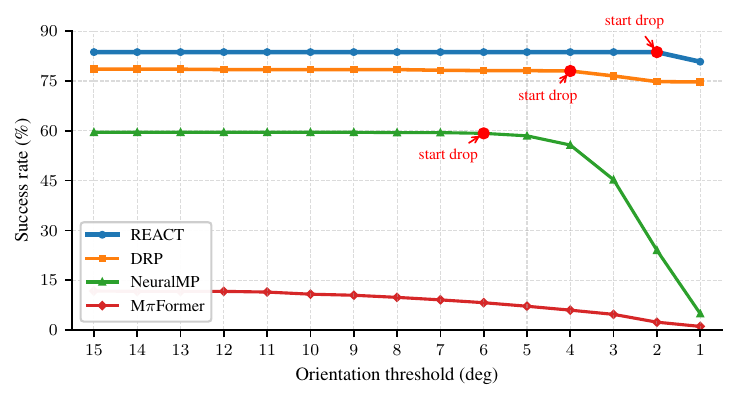}
    \caption{Success rate under different orientation error thresholds. ESCAPE maintains stable performance across a wide range of stricter thresholds and only begins to show a drop when the threshold is tightened below $2^\circ$, indicating stronger goal-reaching precision.}
    \label{fig:orientation_threshold_success}
\end{figure}

Figure~\ref{fig:orientation_threshold_success} further compares the success rates of different methods under progressively stricter orientation thresholds. 
While all methods are evaluated under the same translational criterion, tightening the orientation threshold reveals the relative precision of their terminal goal reaching. 
ESCAPE remains essentially unchanged over a broad range of thresholds, only starting to drop below $2^\circ$, whereas DRP and the other baselines degrade earlier and more significantly. 
This result indicates that ESCAPE achieves stronger goal-reaching precision in addition to collision-free execution, and that its success under the standard $15^\circ$ threshold is not merely due to a loose orientation tolerance.

\section{Baseline Comparison Details}

We compare ESCAPE against classical planners and learning-based motion
policies under the same task definitions, initial states, target poses,
and success criteria. The baselines differ primarily in their observation
assumptions: privileged planners access explicit scene geometry, global
observation policies use global or externally reconstructed point clouds,
and partial-observation policies operate from limited camera or
body-mounted sensing.

\textbf{Classical planning baselines.}
\begin{itemize}
    \item \textbf{AIT*~\cite{strub2020adaptively}}: A sampling-based planner that searches for collision-free paths with access to the full scene geometry.
    \item \textbf{cuRobo~\cite{sundaralingam2023curobo}}: A GPU-accelerated optimization-based planner that uses privileged geometric scene information.
    \item \textbf{cuRobo-Vox~\cite{sundaralingam2023curobo}}: A cuRobo-based planner that replaces precise obstacle models with voxelized inputs derived from point-cloud observations, making it more suitable for unstructured real-world scenes.
\end{itemize}

\textbf{Learning-based baselines.}
\begin{itemize}
    \item \textbf{M$\pi$Nets~\cite{fishman2023motion}}: A neural motion policy that maps point-cloud observations, robot states, and goal conditions to joint-space actions.
    \item \textbf{M$\pi$Former~\cite{pmlr-v270-fishman25a}}: A Transformer-based extension of M$\pi$Nets with teacher-student fine-tuning and hard negative mining. We report the Cubby and Tabletop variants as M$\pi$F-C and M$\pi$F-T, respectively.We evaluate the better-performing Tabletop variant, denoted as M$\pi$F-T.
    \item \textbf{NeuralMP~\cite{dalal2024neural}}: A neural motion policy trained on large-scale generated scenes and improved at evaluation time through test-time optimization (TTO).
    \item \textbf{DRP~\cite{yang2025deep}}: A reactive manipulator motion policy. We evaluate two observation settings: DRP-G uses global point clouds, while DRP-Cam uses iterative external-camera observations.
\end{itemize}

ESCAPE differs from these baselines in that it uses only robot-mounted
near-field depth sensing, the current robot state, and the goal condition,
without external cameras or privileged scene geometry.

\section{Ablation Study on Collision Rollback Fine-tuning Steps}
\label{sec:rollback_ablation}
We compare four fine-tuning strategies under the DAgger framework: 
(i) \texttt{Reset-to-initial} (a baseline strategy that resets to the start configuration upon collision to replan a new trajectory for training); 
(ii) \texttt{rollback15} (rolls back 15 time steps from the collision point); 
(iii) \texttt{rollback20} (rolls back 20 time steps); 
(iv) \texttt{Mixed (15/20/reset)} (randomly chooses among rollback15, rollback20, or reset-to-initial with equal probability). 
Table~\ref{tab:rollback_ablation} reports Goal Reaching (GR), Collision-Free (CF) and Success Rate (SR) for each strategy on Easy, Hard, and Total splits.

\begin{table}[htbp]
\centering
\caption{Comparison of different collision rollback fine-tuning strategies.}
\label{tab:rollback_ablation}
\begin{tabularx}{0.97\textwidth}{@{} l *{9}{>{\centering\arraybackslash}X} @{}}
\toprule
\multirow{2}{*}{Strategy} & \multicolumn{3}{c}{Easy} & \multicolumn{3}{c}{Hard} & \multicolumn{3}{c}{Total} \\
\cmidrule(lr){2-4} \cmidrule(lr){5-7} \cmidrule(lr){8-10}
& GR & CF & SR & GR & CF & SR & GR & CF & SR \\
\midrule
rollback15          & 93.33 & 96.57 & 91.92 & 81.45 & 78.28 & 72.85 & 87.73 & 87.94 & 82.92 \\
rollback20          & 91.11 & 94.34 & 89.09 & \textbf{84.16} & 76.24 & 72.40 & 87.83 & 85.80 & 81.21 \\
Reset-to-initial    & \textbf{95.76} & 94.95 & \textbf{92.73} & 78.73 & 71.04 & 67.87 & 87.73 & 83.67 & 81.00 \\
Mixed (15/20/reset) & 92.93 & \textbf{96.36} & 90.30 & \textbf{84.16} & \textbf{80.09} & \textbf{76.24} & \textbf{88.79} & \textbf{88.69} & \textbf{83.67} \\
\bottomrule
\end{tabularx}
\end{table}

The \texttt{Reset-to-initial} strategy performs best on Easy tasks but struggles in Hard scenarios (SR 67.87\%), confirming that resetting the entire episode prevents the policy from learning how to locally recover from near-collision states in cluttered environments. As shown in Table~\ref{tab:rollback_ablation}, while fixed-length rollback strategies provide marginal gains, they often involve a trade-off between collision avoidance and goal reaching; for instance, \texttt{rollback20} improves goal-reaching accuracy in Hard tasks but leads to over-conservative behavior that hampers overall success. In contrast, the \texttt{Mixed (15/20/reset)} strategy achieves the highest overall success rate. By exposing the policy to a diverse distribution of recovery behaviors—ranging from short-horizon corrections to full-episode restarts—the mixed strategy effectively mitigates the conflict between safety and goal convergence, resulting in superior robustness in complex, high-clutter deployment environments.
\section{Ablation on Collision Inference Steps}
\label{sec:ablation_steps}

To investigate the impact of the temporal unrolling horizon during inference, we conduct an ablation study on the number of numerical simulation steps $n$ used to compute task-level physical gradients. We ablate $n$ from 1 to 5 steps, evaluating performance using Success Rate (SR), Goal Reaching (GR), and Collision-Free (CF) metrics (measured in \%).

\begin{table}[h]
    \centering
    \caption{Performance comparison with varying collision inference steps ($n$).}
    \label{tab:inference_steps}
    \begin{tabular}{lccccc}
        \toprule
        \textbf{Metric (\%)} & \textbf{$n=1$} & \textbf{$n=2$} & \textbf{$n=3$} & \textbf{$n=4$} & \textbf{$n=5$} \\
        \midrule
        \textbf{Success Rate (SR)} & 59.67 & 70.65 & \textbf{78.01} & 76.52 & 39.17 \\
        \textbf{Goal Reaching (GR)} & 70.86 & 78.55 & \textbf{83.99} & 80.47 & 40.23 \\
        \textbf{Collision-Free (CF)} & 69.37 & 78.55 & 82.71 & 88.58 & \textbf{91.04} \\
        \bottomrule
    \end{tabular}
\end{table}

As shown in Table~\ref{tab:inference_steps}, increasing the training inference steps from 1 to 3 improves Success Rate and Goal Reaching Rate by enabling iterative error correction via the differentiable ODE solver. However, further increasing to \(n=5\) leads to a dramatic collapse: Success Rate drops to 39.17\%, Goal Reaching Rate to 40.23\%, while Collision-Free Rate remains high (91.04\%). This is due to error amplification and exploding collision loss gradients, which prevent convergence to valid goal-reaching trajectories; the robot becomes overcautious, often halting or retreating. Thus, a balance between error correction and divergence is required. Our experiments show that \(n=3\) achieves the best performance, consistent with our training unrolling budget.

\section{Detailed Model Parameters}
We list key hyperparameters for training ESCAPE in Table \ref{tab:training_config}.

\begin{table}[htbp]
  \centering
  \caption{Policy Architecture and Training Hyperparameters}
  \label{tab:training_config}
  \begin{tabular}{ll}
    \toprule
    \textbf{Hyperparameter} & \textbf{Value} \\
    \midrule
    \multicolumn{2}{c}{\textbf{Input Parameters}} \\
    \midrule
    Number of ToF sensors & $15$ \\
    ToF sensor resolution & $8 \times 8$ \\
    Raw ToF returns per frame & $15 \times 8 \times 8 = 960$ \\
    Policy point cloud input size & $3072 \times 6$ \\
    Environment / obstacle points & $2048$ \\
    Robot surface points & $1024$ \\
    FPS-sampled encoder points & $128$ \\
    \midrule
    \multicolumn{2}{c}{\textbf{ESCAPE Architecture}} \\
    \midrule
    Current joint state dimension & $7$ \\
    Target joint state dimension & $7$ \\
    Joint state encoder & 2-layer MLP, hidden dims $[64, 64]$ \\
    Goal state encoder & 2-layer MLP, hidden dims $[64, 64]$ \\
    Joint state embedding dim & $64$ \\
    Goal embedding dim & $64$ \\
    Action dimension per step & $7$ \\
    Action chunk shape & $16 \times 7$ \\
    Action chunk horizon ($H$) & $16$ \\
    ODE solver inference steps & $10$ \\
    \midrule
    \multicolumn{2}{c}{\textbf{ESCAPE Training Parameters}} \\
    \midrule
    Optimizer & AdamW \\
    Initial learning rate & $1 \times 10^{-4}$ \\
    Weight decay & $1 \times 10^{-3}$ \\
    Optimizer momentum ($\beta_1, \beta_2$) & $[0.9, 0.95]$ \\
    Batch size per GPU & $128$ \\
    Effective batch size & $1024$ \\
    Warmup steps & $500$ \\
    Learning rate schedule & Cosine decay \\
    Gradient accumulation steps & $1$ \\
    Flow matching vs. consistency ratio & $0.75 : 0.25$ \\
    Pre-training checkpoint & epoch $12$ \\
    GPUs & $8 \times$ NVIDIA A100 \\
    Wall-clock time & 1 day \\
    \midrule
    \multicolumn{2}{c}{\textbf{ESCAPE Loss \& Fine-tuning Parameters}} \\
    \midrule
    Collision safety margin ($m$) & $0.03 \, \mathrm{m}$ \\
    Collision loss weight & $0.8$ \\
    Terminal goal-reaching window ($K$) & $15$ steps \\
    Terminal goal-reaching loss weight & $0.1$ \\
    GPUs & $8 \times$ NVIDIA A100 \\
    Wall-clock time & 3 hours \\
    \bottomrule
  \end{tabular}
\end{table}

\end{document}